%% file: root.tex
\documentclass[letterpaper, 10 pt, conference]{ieeeconf}  

\IEEEoverridecommandlockouts                              

\usepackage{booktabs}
\usepackage{multirow}
\usepackage{graphicx}
\usepackage{xcolor}
\usepackage{amsmath}
\usepackage{subcaption}
\usepackage{dblfloatfix}
\usepackage{url}
\usepackage{amssymb}

\title{\LARGE \bf
EagleVision: A Multi-Task Benchmark for Cross-Domain Perception in High-Speed Autonomous Racing
}

\author{
Zakhar Yagudin$^{1, 2}$, Murad Mebrahtu$^{2}$, Ren Jin$^{3}$, Jiaqi Huang$^{3}$, \\ Yujia Yue$^{3}$, Dzmitry Tsetserukou$^{1}$, Jorge Dias$^{2}$, Majid Khonji$^{2}$
\thanks{$^{1}$Skolkovo Institute of Science, Intelligent Space Robotics Laboratory, Center for Engineering Systems and Sciences}%
\thanks{$^{2}$Khalifa University, KUCARS-KU Center for Autonomous Robotic Systems, Department of Computer Science}%
\thanks{$^{3}$Beijing Institute of Technology, Beijing Key Laboratory of UAV Autonomous Control}%
}

\begin{document}

\maketitle
\thispagestyle{empty}
\pagestyle{empty}

\begin{abstract}
High-speed autonomous racing presents extreme perception challenges, including large relative velocities and substantial domain shifts from conventional urban-driving datasets. Existing benchmarks do not adequately capture these high-dynamic conditions. We introduce EagleVision, a unified LiDAR-based multi-task benchmark for 3D detection and trajectory prediction in high-speed racing, providing newly annotated 3D bounding boxes for the Indy Autonomous Challenge dataset (14,893 frames) and the A2RL Real competition dataset (1,163 frames), together with 12,000 simulator-generated annotated frames, all standardized under a common evaluation protocol. Using a dataset-centric transfer framework, we quantify cross-domain generalization across urban, simulator, and real racing domains. Urban pretraining improves detection over scratch training (NDS 0.72 vs. 0.69), while intermediate pretraining on real racing data achieves the best transfer to A2RL (NDS 0.726), outperforming simulator-only adaptation. For trajectory prediction, Indy-trained models surpass in-domain A2RL training on A2RL test sequences (FDE 0.947 vs. 1.250), highlighting the role of motion-distribution coverage in cross-domain forecasting. EagleVision enables systematic study of perception generalization under extreme high-speed dynamics. The dataset and benchmark are publicly available at \url{https://avlab.io/EagleVision}.
\end{abstract}

\input{sections/1_intro}
\input{sections/2_dataset}
\input{sections/3_benchmark}
\input{sections/4_baselines}
\input{sections/5_results}
\input{sections/6_conclusion}

\bibliographystyle{IEEEtran}

\bibliography{references}

\end{document}

%% file: sections/1_intro.tex
\section{Introduction}
\subsection{High-Speed Racing Perception Challenges}
\begin{figure}[t]
\centering
\includegraphics[width=\columnwidth]{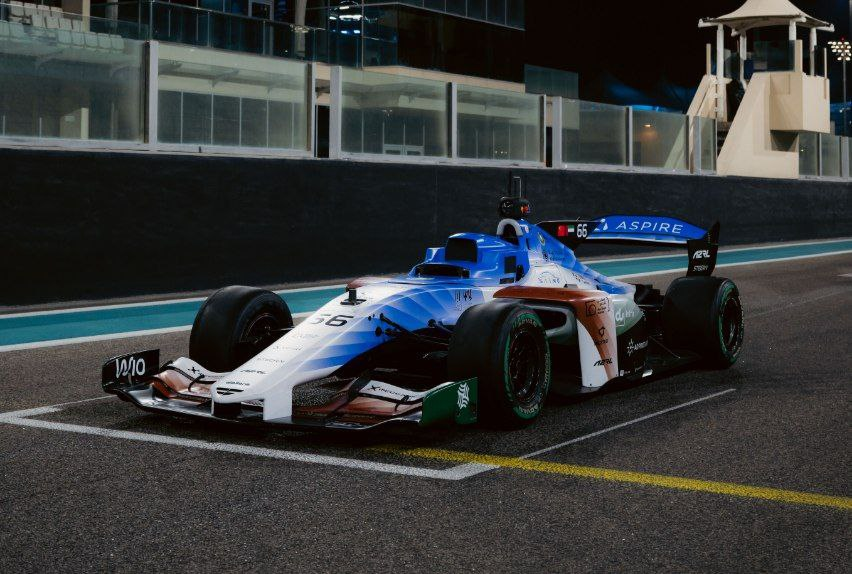}
\caption{Autonomous racing platform operating under real A2RL competition.}
\label{fig:platform}
\end{figure}

In recent years, perception technologies for autonomous driving have achieved remarkable progress in conventional traffic scenarios such as urban roads and highways. Benefiting from large-scale public datasets, including KITTI~\cite{geiger2013vision}, nuScenes~\cite{caesar2020nuscenes}, Waymo Open Dataset~\cite{sun2020scalability}, and Argoverse 2~\cite{wilson2023argoverse}, significant advances have been made in tasks such as object detection, multi-object tracking, and trajectory prediction, leading to substantial improvements in both algorithmic accuracy and system robustness. However, these datasets are primarily designed for low- to medium-speed, structured traffic environments, whose data acquisition conditions, dynamic ranges, and interaction complexities differ fundamentally from those encountered in extreme high-speed driving scenarios.
Figure~\ref{fig:platform} illustrates the autonomous racing platform used for data collection during A2RL competitions. The extreme speed regime and sparse track environment highlight the domain shift relative to conventional urban driving datasets.
The A2RL Real dataset was collected during the official Abu Dhabi Autonomous Racing League (A2RL) competition, where our system achieved 2nd place in the Silver Race category, ensuring genuine high-speed competitive racing conditions.

Autonomous racing represents highly challenging operating conditions for perception systems. Compared to urban traffic, vehicles operate at significantly higher speeds, resulting in large relative velocities, aggressive accelerations, and rapid motion changes. Consequently, sensor observations are affected by motion blur, scale variations, and abrupt viewpoint changes. The racing environment also exhibits non-uniform object distributions and strong interactions between vehicles, leading to increased dynamic uncertainty. These factors cause perception models trained on conventional road-driving datasets to suffer performance degradation when deployed in racing scenarios, making it difficult to meet strict reliability and latency requirements. Similar performance degradation under distribution shift has been observed in cross-domain 3D object detection studies~\cite{xu2021spg,li2023gpa3d}.

Although increasing attention has recently been paid to autonomous racing and high-speed driving, publicly available perception datasets specifically designed for racing scenarios remain extremely limited. Existing datasets are often constrained by limited scale, narrow task coverage, or a lack of systematic support for real-world high-speed conditions and multi-task perception problems. As a result, they are insufficient for comprehensive evaluation and fair comparison of detection, tracking, and prediction algorithms under high-speed settings. This limitation has, to a certain extent, hindered the development of perception algorithms tailored for extreme driving conditions.

Motivated by these observations, this paper focuses on multi-task perception in high-speed autonomous racing scenarios and presents a dedicated racing perception dataset that supports object detection, multi-object tracking, and trajectory prediction.

\subsection{Existing Datasets for Autonomous Driving and Racing}
Public datasets have played a critical role in advancing perception algorithms for autonomous driving. The KITTI dataset~\cite{geiger2013vision} established early standardized benchmarks for 3D object detection and multi-object tracking. nuScenes~\cite{caesar2020nuscenes} and the Waymo Open Dataset~\cite{sun2020scalability} further introduced large-scale, multi-modal sensor data covering diverse urban traffic scenarios, while Argoverse 2~\cite{wilson2023argoverse} emphasizes trajectory prediction under complex interactions with the support of high-definition maps. Recent trajectory prediction methods include graph-based models such as VectorNet~\cite{gao2020vectornet} and transformer-based approaches such as MTR++~\cite{shi2023mtrplusplus}, which achieve strong performance in structured urban environments. These datasets are highly valuable in terms of data scale, annotation quality, and task coverage. Numerous multi-sensor 3D detection architectures have been developed based on these datasets. For instance, Fadet~\cite{guo2025fadet} introduces a local-feature attention mechanism for LiDAR-camera fusion to enhance 3D object detection performance.
Comprehensive surveys summarize the evolution of LiDAR-based 3D object detection and representation learning in autonomous driving~\cite{qian2022survey3d,alaba2022lidar_survey}. However, such models are primarily evaluated under structured urban traffic conditions, and their robustness under extreme high-speed racing dynamics remains unclear.

In parallel with the development of perception datasets, recent studies have explored high-level scene understanding and decision-making through vision-language models. For example, VLM-auto~\cite{guo2024vlm} leverages large vision-language models to enable human-like behavioral reasoning in complex road scenes. While such approaches demonstrate strong semantic understanding and contextual reasoning capabilities, their performance heavily depends on robust low-level perception modules, whose reliability under extreme high-speed racing conditions remains underexplored.

The RACECAR dataset~\cite{kulkarni2023racecar} is a publicly available dataset specifically focused on high-speed autonomous racing, derived from real competition data collected by multiple teams participating in the Indy Autonomous Challenge. It records multi-modal sensor data at speeds of up to 273 km/h across a variety of racing scenarios, including single-lap runs, multi-vehicle competition, overtaking maneuvers, high-acceleration segments, obstacle avoidance, and pit-lane operations, covering a total of 11 representative track scenarios. The dataset is released in both ROS2 and nuScenes formats to facilitate its adoption by existing autonomous driving frameworks.

RACECAR contributes valuable real-world sensor data for fundamental problems such as localization, object detection, tracking, and mapping under highly dynamic conditions. Moreover, the support for the nuScenes format enables partial alignment with standard autonomous driving benchmarks. Nevertheless, the dataset currently lacks mature and comprehensive annotations (e.g., full 3D bounding box labels) as well as unified evaluation protocols for perception tasks. Consequently, it is mainly used for exploratory studies on basic perception and localization, while systematic evaluation of multi-task perception and prediction in high-speed scenarios remains limited.

The BETTY dataset~\cite{nye2025betty} is a recently proposed large-scale, high-dynamic, multi-modal autonomous racing dataset designed to support full-stack autonomy research, including state estimation, vehicle dynamics modeling, motion prediction, and perception. The dataset contains several hours of data, totaling approximately 32 TB, collected from multiple autonomous racing platforms. It provides rich information spanning perception sensors (cameras, LiDARs, radars), software stack outputs, semantic metadata, and ground-truth labels, and covers diverse racing environments such as high-acceleration corners, feature-sparse tracks, and GPS-denied scenarios. In addition, BETTY captures extreme vehicle behaviors, including loss of control, tire slip, and stability limits, making it well suited for full-stack system training and evaluation.

Despite its scale and broad task coverage, BETTY primarily targets end-to-end training and motion modeling for full-stack autonomy. For research that focuses specifically on perception benchmarks such as visual detection and tracking, the dataset’s annotation structure and task definitions are relatively complex, which poses challenges for standardized evaluation and fair comparison, particularly for single-task benchmarks such as detection and trajectory prediction.
\subsection{Motivation and Dataset-Centric Evaluation}
By providing synchronized sensory input and perception annotations, our dataset enables researchers to systematically evaluate perception algorithms under high-speed racing conditions using a unified benchmark. In contrast to common practice—where detectors trained in conventional road-driving datasets are directly transferred to racing scenarios—we adopt a dataset-centric evaluation methodology that explicitly quantifies the impact of training data distributions.
While prior works explore unsupervised domain adaptation and multi-source training for 3D detection~\cite{tsai2023ms3dplusplus,li2023bev_dg}, systematic analysis under extreme high-speed racing conditions remains limited.

Specifically, we evaluate object detectors trained on multiple existing datasets and their combinations and assess their performance on a unified racing test set. This allows us to isolate how different data sources and dataset compositions affect detection performance in extreme high-speed scenarios. Through this design, researchers can analyze performance degradation caused by domain mismatch, measure the benefits of incorporating race-specific data, and study complementarity and redundancy among datasets. Geometry-aware and sim-to-real adaptation strategies have been shown to mitigate cross-domain gaps in 3D detection~\cite{li2023gpa3d,wozniak2024sim2real}.

Recent end-to-end driving frameworks such as METDrive~\cite{guo2025metdrive} integrate multimodal perception and temporal guidance into unified architectures. While such systems reduce the reliance on modular pipelines, their generalization ability is still fundamentally constrained by the distribution of training data. Therefore, systematic dataset-centric evaluation under high-speed racing conditions is essential for understanding performance limits and domain gaps.
\subsection{Contributions}

In summary, we make the following contributions:

\begin{itemize}

\item \textbf{EagleVision Dataset:} We introduce the first unified LiDAR-based perception benchmark for high-speed autonomous racing with newly provided 3D bounding-box annotations across three domains:
(i) the Indy Autonomous Challenge (IAC) dataset with 14{,}893 manually labeled frames,
(ii) the A2RL Real-World dataset with 1{,}163 manually labeled competition frames,
and (iii) the A2RL Simulator dataset with 12{,}000 ground-truth annotated frames.
All datasets are standardized into a common annotation and coordinate convention using the \textit{SUSTechPOINTS} format~\cite{sustechpoints}, enabling consistent cross-domain evaluation.

\item \textbf{Competition-Grade Real Racing Data:}
The A2RL Real dataset was collected during the official Abu Dhabi Autonomous Racing League (A2RL) competition, where our autonomous racing system achieved 2nd place in the Silver Race category. The dataset therefore reflects genuine high-speed competitive racing conditions rather than controlled laboratory scenarios.

\item \textbf{Dataset-Centric Cross-Domain Evaluation Protocol:}
We propose a systematic transfer-learning framework spanning urban-to-racing, simulator-to-real, and racing-to-racing adaptation. This protocol explicitly quantifies how training data distributions affect perception performance under extreme high-speed dynamics.

\item \textbf{Unified Multi-Task Benchmark:}
We provide standardized tasks, metrics, and baselines for both 3D object detection and trajectory prediction, including cross-vehicle prediction splits and split-difficulty analysis to study generalization under high-speed motion regimes.

\end{itemize}

%% file: sections/2_dataset.tex
\section{Dataset and Annotation Protocol}

This work consolidates high-speed autonomous racing data from the Indy Autonomous Challenge (IAC)~\cite{IAC2023} and the A2RL competition~\cite{A2RL2023} into a unified benchmark for 3D object detection and trajectory prediction. All datasets are standardized under a common annotation format and coordinate convention to enable consistent cross-domain evaluation.
Throughout this paper, we denote the Indy dataset as $\mathcal{I}$, the A2RL Simulator dataset as $\mathcal{S}$, and the A2RL Real-World dataset as $\mathcal{R}$.




\subsection{Indy Autonomous Challenge (IAC)}

The Indy Autonomous Challenge (IAC) vehicles are equipped with a Luminar Iris 360° long-range LiDAR sensor with a maximum range of 275\,m, operating at 20\,Hz. We utilize the publicly available \textit{T-MULTI-FAST-POLI} session, which contains multi-lap high-speed racing data collected from two vehicles, denoted as $\mathcal{I}^{v3}$ and $\mathcal{I}^{v5}$.

The dataset consists of raw ROS bag recordings with GPS-based vehicle state information but does not include 3D object detection annotations. Consequently, all 3D bounding boxes were manually annotated from LiDAR point clouds.

In total, 14,893 frames were labeled. Typical scenes contain a single opponent vehicle per frame under high-speed racing conditions.
\subsection{A2RL Simulator Dataset}

Synthetic data was collected using the official A2RL racing simulator. The simulated platform replicates the perception configuration of the competition vehicle and allows spawning multiple racing agents.

We generated 12,000 annotated frames from the simulator.  
Each frame contains between 4 to 7 vehicles on average, enabling multi-agent interaction scenarios.

Ground-truth 3D bounding boxes were directly obtained from the simulator engine and exported into the same annotation format used for IAC.

\subsection{A2RL Real-World Dataset}

The A2RL competition vehicle is equipped with three Seyond Falcon Ultra Long-Range LiDAR sensors: one forward-facing unit and two side-mounted units oriented rearward.

Real-world racing data was recorded as ROS bags and subsequently exported for annotation.  
A total of 1,163 frames were manually labeled.  
Each frame typically contains a single opponent vehicle.

Compared to simulator data, the real dataset exhibits sensor noise, reflectivity variance, and partial occlusions during overtaking maneuvers.

\subsection{Annotation Format and Coordinate Convention}

All datasets are unified under a consistent Position-Scale-Rotation (PSR) JSON schema.  
Each object is represented as:

Each annotated object is represented by a 3D bounding box parameterized by its center position $(x,y,z)$, dimensions $(l,w,h)$, and yaw rotation around the vertical axis. All annotations are provided in the ego-vehicle coordinate frame.
All bounding boxes are defined in the LiDAR coordinate frame without motion compensation to ensure consistency across real and synthetic domains.

Only a single semantic class (\texttt{Car}) is considered in this benchmark.

\subsection{Prediction Annotations}

For trajectory-prediction benchmarking, frame-level annotations are provided in a single file per dataset split and domain ($\mathcal{I}_{v3}$, $\mathcal{I}_{v5}$, and $\mathcal{R}$). Each entry corresponds to a single frame and contains:

\begin{itemize}
    \item a unique frame identifier (\texttt{frame\_id}),
    \item a timestamp,
    \item the 3D vehicle position (\texttt{x}, \texttt{y}, \texttt{z}),
    \item the vehicle orientation represented as a unit quaternion (\texttt{q\_x}, \texttt{q\_y}, \texttt{q\_z}, \texttt{q\_w}).
\end{itemize}

Trajectories are constructed from consecutive frames using fixed-length observation and prediction windows. 



\subsection{Dataset Statistics}

\begin{table}[t]
\centering
\caption{Summary of racing datasets used for the detection benchmark.}
\label{tab:dataset_stats}
\resizebox{\columnwidth}{!}{
\begin{tabular}{lccc}
\toprule
 & IAC & A2RL Sim & A2RL Real \\
\midrule
LiDAR sensor & 3$\times$ Luminar Iris & Virtual LiDAR & 3$\times$ Seyond Falcon \\
Frequency (Hz) & 20 & 10 & 10 \\
Annotated frames & 14,893 & 12,000 & 1,163 \\
Avg. objects/frame & 1 & 4 & 1 \\
Points per scan (approx.) & $\sim$67k & $\sim$144k & $\sim$100k \\
\bottomrule
\end{tabular}
}
\end{table}

Table~\ref{tab:dataset_stats} highlights the key structural differences across domains, including point cloud density, object multiplicity, and acquisition frequency. While IAC and A2RL Real primarily contain single-opponent scenarios, the simulator provides higher object density and perfectly aligned ground-truth annotations, enabling controlled multi-agent experiments. A2RL Real consists of 1,163 annotated frames. For detection experiments, the dataset is divided into training and validation subsets, with approximately 929 frames used for training and the remaining frames reserved for validation.

%% file: sections/3_benchmark.tex
\section{Benchmark Tasks and Evaluation Protocol}
\label{sec:protocol}

\begin{figure}
\centering

\begin{subfigure}{\columnwidth}
    \centering
    \includegraphics[width=\columnwidth]{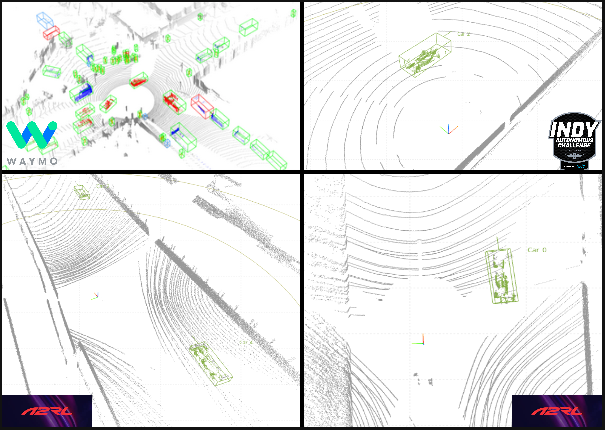}
    \caption{Detection datasets and representative LiDAR annotations.}
\end{subfigure}

\vspace{0.5em}

\begin{subfigure}{\columnwidth}
    \centering
    \includegraphics[width=\columnwidth]{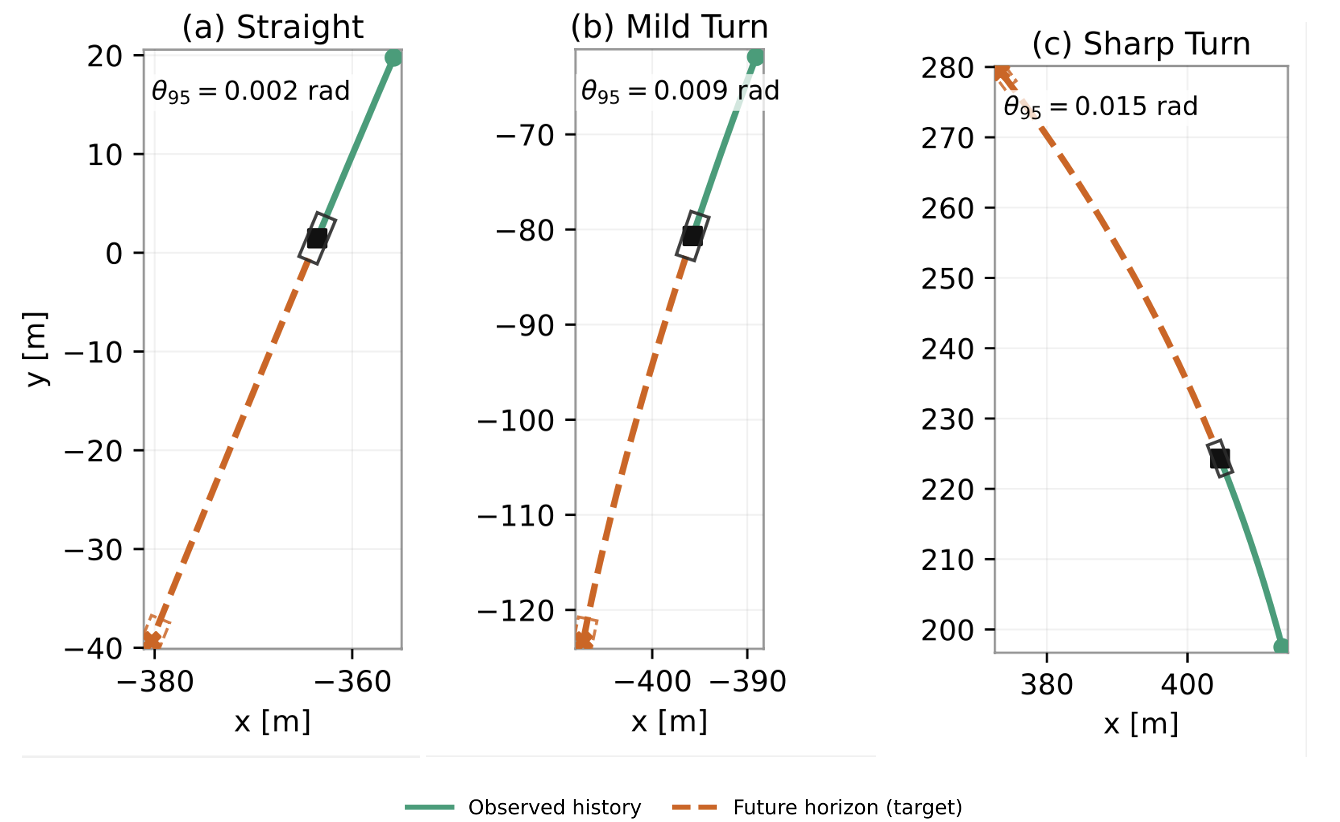}
    \caption{Sample trajectories from Prediction annotations}
\end{subfigure}

\caption{Overview of the proposed benchmark. 
(a) 3D detection across urban, simulator, and real racing domains. 
(b) Trajectory prediction settings, illustrating straight, mild turn, and sharp turn scenarios based on ground-truth annotations.}
\label{fig:overview}
\end{figure}

This section defines the detection and trajectory prediction tasks, evaluation metrics, and experimental configurations used throughout the paper. Figure~\ref{fig:overview} provides an overview of the EagleVision benchmark design. The benchmark unifies detection and trajectory prediction tasks across multiple domains, including urban driving, simulator-based racing, and real-world high-speed racing.

\subsection{3D Detection Task}

The 3D detection task consists of localizing racing vehicles in the LiDAR coordinate frame using oriented 3D bounding boxes.  
Only a single semantic class (\texttt{Car}) is considered.

Detection is performed within the spatial range:
\[
[-60, 60] \text{ m (x)}, \quad [-60, 60] \text{ m (y)}, \quad [-2, 4] \text{ m (z)}.
\]

All predictions are evaluated against manually annotated ground-truth boxes defined in Section~II.

\subsection{Detection Metrics}

We adopt a distance-based evaluation protocol inspired by the nuScenes detection benchmark~\cite{caesar2020nuscenes}.  
Predicted boxes are matched to ground-truth boxes based on center-distance thresholds.

The following metrics are reported:

\begin{itemize}
    \item \textbf{AP:} Average Precision computed over center-distance thresholds 
    $\{0.25, 0.5, 1.0, 2.0\}$ meters.
    
    \item \textbf{ATE:} Average Translation Error (center distance).
    
    \item \textbf{ASE:} Average Scale Error.
    
    \item \textbf{AOE:} Average Orientation Error (yaw).
\end{itemize}

We report a reduced variant of the nuScenes Detection Score (NDS)~\cite{caesar2020nuscenes}, computed as:

\[
\text{NDS} = \frac{1}{10} \left( 5 \cdot AP + 3 - ATE - ASE - AOE \right).
\]

Since velocity and attribute annotations are not available in our dataset, velocity (AVE) and attribute (AAE) components from the original nuScenes formulation are omitted.
\subsection{Trajectory Prediction Task}
\input{sections/3.2_trajectory_prediction_task}

\subsection{Detection Transfer Protocol}

We denote A2RL Real as $\mathcal{R}$, Indy as $\mathcal{I}$, Simulator as $\mathcal{S}$, and Waymo pretraining as $\mathcal{W}$~\cite{Sun2020Waymo}.

The evaluated transfer strategies are:

\begin{itemize}

\item \textbf{$\mathcal{R}$ (Scratch):}  
Training on $\mathcal{R}$ from random initialization.

\item \textbf{$\mathcal{W} \rightarrow \mathcal{R}$:}  
Waymo-pretrained model finetuned on $\mathcal{R}$.

\item \textbf{$\mathcal{W} \rightarrow (\mathcal{R}+\mathcal{S})_{1:1}$:}  
Joint finetuning on $\mathcal{R}$ and $\mathcal{S}$ with equal sampling ratio.

\item \textbf{$\mathcal{W} \rightarrow \mathcal{S}_{10} \rightarrow \mathcal{R}$:}  
Pretraining on $\mathcal{S}$ for 10 epochs, followed by finetuning on $\mathcal{R}$.

\item \textbf{$\mathcal{W} \rightarrow (\mathcal{R}+0.1\mathcal{S})$:}  
Finetuning on $\mathcal{R}$ augmented with 10\% simulator samples.

\item \textbf{$\mathcal{W} \rightarrow (\mathcal{R}+0.1\mathcal{I})$:}  
Finetuning on $\mathcal{R}$ augmented with 10\% Indy samples.

\item \textbf{$\mathcal{W} \rightarrow \mathcal{I}_{10} \rightarrow \mathcal{R}$:}  
Pretraining on $\mathcal{I}$ for 10 epochs before finetuning.

\item \textbf{$\mathcal{W} \rightarrow (\mathcal{S}+\mathcal{I})_{30} \rightarrow \mathcal{R}$:}  
Joint pretraining on $\mathcal{S}$ and $\mathcal{I}$ for 30 epochs.

\item \textbf{$\mathcal{W} \rightarrow \mathcal{I}_{5} \rightarrow \mathcal{S}_{8} \rightarrow \mathcal{R}$:}  
Two-stage pretraining before final finetuning.
\end{itemize}

For all experiments, the best checkpoint is selected according to validation NDS.

%% file: sections/3.2_trajectory_prediction_task.tex

The trajectory prediction task aims to estimate the future motion of detected vehicles based on past observations. 
Given a sequence of historical positions expressed in a global reference frame, the model predicts future positions over a fixed time horizon.

Vehicles in racing scenarios operate at high speeds (approximately 50\,m/s in $\mathcal{I}$), leading to rapidly growing prediction uncertainty over longer horizons. 
We therefore adopt a prediction horizon of 1\,s with 0.5\,s of historical observations as input, providing a balance between contextual information and prediction stability.

The dataset operates at approximately 20\,Hz. 
To ensure consistent temporal spacing for reliable motion derivative estimation, all trajectories are resampled to a fixed rate of 20\,Hz. 
This removes temporal jitter and irregular frame spacing observed in the raw recordings.

For $\mathcal{I}$, vehicle trajectories are derived directly from global odometry measurements. 
For $\mathcal{R}$, target vehicle positions originate from detection annotations expressed in the ego-vehicle frame. 
All bounding box centers are transformed into the global map frame using the per-frame ego pose to remove ego-motion effects.

To mitigate residual temporal jitter and spurious position jumps, a short-horizon constant-velocity smoothing step is applied in the global frame. 
This improves trajectory continuity while preserving the underlying high-speed motion characteristics.

The resulting trajectories are globally referenced and temporally consistent, enabling stable learning of motion prediction models.

\paragraph{Prediction Metrics and Evaluation Protocol.}
We evaluate trajectory prediction performance using standard displacement-based metrics: Average Displacement Error (ADE) and Final Displacement Error (FDE). 
Let $\mathbf{x}_{t}^{(i)} \in \mathbb{R}^2$ denote the ground-truth position of agent $i$ at prediction step $t$ in the global/map frame, and $\hat{\mathbf{x}}_{t}^{(i)}$ the corresponding prediction.  
For a prediction horizon of \(T\) steps and \(N\) predicted trajectories, ADE is defined as
\[
\mathrm{ADE} = \frac{1}{N T} \sum_{i=1}^{N} \sum_{t=1}^{T} \left\| \hat{\mathbf{x}}_{t}^{(i)} - \mathbf{x}_{t}^{(i)} \right\|_2,
\]
and FDE is defined as:
\[
\mathrm{FDE} = \frac{1}{N} \sum_{i=1}^{N} \left\| \hat{\mathbf{x}}_{T}^{(i)} - \mathbf{x}_{T}^{(i)} \right\|_2,
\]
where, \(\hat{\mathbf{x}}_{t}^{(i)} \in \mathbb{R}^2\) and \(\mathbf{x}_{t}^{(i)} \in \mathbb{R}^2\) denote the predicted and ground-truth positions, respectively, for trajectory \(i\) at time step \(t\). 
ADE quantifies the mean displacement error over the full prediction horizon, whereas FDE quantifies the displacement error at the final prediction step.





\subsection{Prediction Experimental Setups}

\input{sections/3_2_Prediction_setup}

%% file: sections/3_2_Prediction_setup.tex






The prediction model is trained using the $\mathcal{I}$ and $\mathcal{R}$ datasets. 
The availability of two vehicles within $\mathcal{I}$ enables evaluation under both intra-domain and cross-vehicle configurations.

To evaluate model stability, cross-vehicle generalization, and domain adaptation capability, the following experimental setups are defined:



\begin{enumerate}
    \item \textbf{Combined-Vehicle Split (Intra-domain Evaluation):}  
    The complete Indy dataset, $\mathcal{I} = \mathcal{I}^{v3} \cup \mathcal{I}^{v5}$, is divided into training, validation, and test sets using a 70\% / 15\% / 15\% split.

    \item \textbf{Cross-Vehicle Setup I ($\mathcal{I}^{v3} \rightarrow \mathcal{I}^{v5}$):}  
    $\mathcal{I}^{v3}$ is divided into training and validation sets using an 80\% / 20\% split, while $\mathcal{I}^{v5}$ is reserved exclusively for testing.

    \item \textbf{Cross-Vehicle Setup II ($\mathcal{I}^{v5} \rightarrow \mathcal{I}^{v3}$):}  
    $\mathcal{I}^{v5}$ is divided into training and validation sets using an 80\% / 20\% split, while $\mathcal{I}^{v3}$ is used exclusively for testing.

    \item \textbf{$\mathcal{R}$-Only Training (Intra-domain Evaluation):}  
    The dataset $\mathcal{R}$ is divided into training, validation, and test sets using a 70\% / 15\% / 15\% split.
\end{enumerate}

In all $\mathcal{I}$-based setups, the trained models are additionally evaluated on $\mathcal{R}$ to assess cross-domain generalization and domain adaptation performance.
\paragraph{Dataset Statistics.}
Under the current preprocessing setup ($20\,\mathrm{Hz}$, observation length $=10$, prediction length $=20$, frame gap $=1$), the resulting trajectory counts are summarized in Table~\ref{tab:cross_split_stats}.

\paragraph{Cross-Domain Split Statistics.}
For cross-domain experiments, we train on one $\mathcal{I}$ domain and evaluate on two test domains. 
Specifically, for $\mathcal{I}^{v3}$ training, the in-domain test is $\mathcal{I}^{v5}$; for $\mathcal{I}^{v5}$ training, the in-domain test is $\mathcal{I}^{v3}$. 
For the $\mathcal{I}$ setup (both vehicles used in training), the in-domain test corresponds to the held-out $\mathcal{I}$ test split. 
In all cases, we additionally report out-of-domain performance on the $\mathcal{R}$ test split.

\begin{table}[h]
\centering
\footnotesize
\setlength{\tabcolsep}{5pt}
\begin{tabular}{lrrrr}
\hline
Setup & Train & Validation & $\mathcal{I}$ Test & $\mathcal{R}$ Test \\
\hline
$\mathcal{I}$ & $21{,}848$ & $4{,}611$ & $4{,}529$ & $328$ \\
$\mathcal{I}^{v3}$ & $12{,}259$ & $2{,}959$ & $15{,}828$ & $328$ \\
$\mathcal{I}^{v5}$ & $12{,}668$ & $3{,}130$ & $15{,}247$ & $328$ \\
$\mathcal{R}$ (in-domain) & $212$ & $29$ & $-$ & $29$ \\
\hline
\end{tabular}
\caption{Trajectory counts for cross-domain and $\mathcal{R}$ in-domain splits.}
\label{tab:cross_split_stats}
\end{table}

%% file: sections/4_baselines.tex
\section{Baselines}
\label{sec:baselines}

\subsection{Detection Baseline}

We adopt a PointPillars-style voxel-based 3D detector~\cite{Lang2019PointPillars} with a CenterPoint detection head~\cite{Yin2021CenterPoint}.  
Unless otherwise specified, models are initialized from a Waymo-pretrained checkpoint~\cite{Sun2020Waymo}.

\paragraph{Input Representation}
Point clouds are voxelized with a voxel size of $(0.24, 0.24, 6.0)$\,m within a spatial range of
$[-60, 60]$\,m (x, y) and $[-2, 4]$\,m (z).
A single LiDAR sweep is used per frame.

\paragraph{Network Architecture}
The backbone follows the standard PointPillars architecture with a pillar feature encoder and RPN-style neck.  
The detection head predicts center offsets, box dimensions, height, and orientation.

\paragraph{Training Setup}
Models are trained for 50 epochs using the Adam optimizer with weight decay $0.01$.  
A one-cycle learning rate schedule is employed with a maximum learning rate of $3\times10^{-3}$.  
Global rotation and scale augmentations are applied during training.

\paragraph{Inference}
Non-maximum suppression (NMS) is applied with IoU threshold $0.2$, pre-NMS limit of 4096 boxes, and post-NMS limit of 500 boxes.  
The confidence threshold is set to $0.3$.

For all transfer strategies defined in Section~III, the best checkpoint is selected according to validation NDS.

\subsection{Trajectory Prediction Baseline}


We adopt a sequence-to-sequence LSTM baseline for trajectory forecasting. 
Given an observed trajectory segment expressed in the global map frame, the model predicts future motion over a fixed time horizon. 
An LSTM architecture is selected due to its effectiveness in modeling short-term temporal dependencies in sequential motion data, while remaining computationally efficient and robust under moderate training data sizes. 
Since the current benchmark relies solely on past trajectory information without additional contextual inputs, a lightweight recurrent model provides a strong and stable baseline.

\paragraph{Model Architecture}
The implemented baseline is an encoder--decoder LSTM model. 
For each sample, we consider the sequence of observed positions, where the position at time step $t$ is denoted by:
\[
\mathbf{p}_t = (x_t, y_t)
\]
The corresponding planar displacements are defined as:
\[
\mathbf{d}_t = \mathbf{p}_t - \mathbf{p}_{t-1}
\]
The encoder processes the history of observed displacements, and the decoder autoregressively predicts future displacements.
A linear layer maps the decoder hidden states to 2D displacement outputs at each future time step.

The predicted future positions are reconstructed from the last observed position $\mathbf{p}_{\mathrm{obs}}$ as:
\[
\hat{\mathbf{p}}_1 = \mathbf{p}_{\mathrm{obs}} + \hat{\mathbf{d}}_1,
\qquad
\hat{\mathbf{p}}_t = \hat{\mathbf{p}}_{t-1} + \hat{\mathbf{d}}_t, \quad t \ge 2 
\]
\paragraph{Temporal Window}
We use $10$ observed steps and predict $20$ future steps (i.e., $0.5\,\mathrm{s}$ history and $1.0\,\mathrm{s}$ prediction at $20\,\mathrm{Hz}$).

\paragraph{Loss Function}
Training minimizes MSE in \emph{displacement space}:
\[
\mathcal{L}_{\mathrm{disp}} = \frac{1}{T}\sum_{t=1}^{T}\left\|\hat{\mathbf{d}}_t - \mathbf{d}_t\right\|_2^2,
\]
where \(T\) is the prediction horizon, \(\mathbf{d}_t \in \mathbb{R}^2\) is the ground-truth planar displacement at step \(t\), and \(\hat{\mathbf{d}}_t \in \mathbb{R}^2\) is the predicted displacement.
The term \(\left\|\hat{\mathbf{d}}_t - \mathbf{d}_t\right\|_2^2\) denotes the squared Euclidean error at each step, and the loss averages this error over all future steps.
ADE/FDE are then computed in \emph{position space} after integrating predicted displacements.
\paragraph{Training Protocol}
We ran a 16-configuration sweep per dataset setup:
\begin{itemize}
    \item LSTM layers: $\{2,4\}$
    \item Hidden size: $\{64,128\}$
    \item Learning rate: $\{10^{-3}, 5\times10^{-4}\}$
    \item Batch size: $\{16,32\}$
\end{itemize}
with $64$ epochs per run (and patience-based early stopping on validation \emph{loss}). 
Optimization uses Adam, with gradient clipping.

\paragraph{Evaluation Protocol}
For each trained configuration, we report Average Displacement Error (ADE) and Final Displacement Error (FDE) on the train, validation, and in-domain test splits. 
For cross-domain analysis, we additionally evaluate on the A2RL test split and report ADE and FDE on that split.

\paragraph{Model Selection For Comparison}
All reported ADE/FDE pairs are taken from the same trained configuration (no metric mixing across runs). 
For hyperparameter-sweep comparison, configurations are ranked using in-domain test performance, while A2RL test metrics are used only to assess out-of-domain generalization and are not used for model selection.


%% file: sections/5_results.tex
\section{Experimental Results}
\label{sec:results}

This section presents quantitative detection and trajectory prediction results, followed by an analysis of cross-domain transfer behavior.

\subsection{Detection Results}

Table~\ref{tab:detection_results} reports detection performance on A2RL Real under all transfer strategies defined in Section~III.

\begin{table}[t]
\centering
\caption{Detection performance on A2RL Real.}
\label{tab:detection_results}
\begin{tabular*}{\columnwidth}{@{\extracolsep{\fill}}lccccc@{}}
\toprule
Setup & AP $\uparrow$ & ATE $\downarrow$ & ASE $\downarrow$ & AOE $\downarrow$ & NDS $\uparrow$ \\
\midrule
$\mathcal{R}$ 
& 0.843 
& 0.1796 
& 0.0716 
& 0.0372 
& 0.69266 \\

$\mathcal{W} \rightarrow \mathcal{R}$ 
& 0.890 
& 0.1702 
& 0.0333 
& 0.0231 
& 0.72234 \\

$\mathcal{W} \rightarrow (\mathcal{R}+\mathcal{S})_{1:1}$ 
& 0.847 
& 0.1918 
& 0.0348 
& 0.0254 
& 0.69830 \\

$\mathcal{W} \rightarrow \mathcal{S}_{10} \rightarrow \mathcal{R}$ 
& 0.879 
& 0.1580 
& 0.0294 
& 0.0208 
& 0.71868 \\

$\mathcal{W} \rightarrow (\mathcal{R}+0.1\mathcal{S})$ 
& 0.873 
& 0.1589 
& 0.0310 
& 0.0209 
& 0.71542 \\

$\mathcal{W} \rightarrow (\mathcal{R}+0.1\mathcal{I})$ 
& 0.883 
& 0.1666 
& 0.0413 
& 0.0213 
& 0.71858 \\

$\mathcal{W} \rightarrow \mathcal{I}_{10} \rightarrow \mathcal{R}$ 
& \textbf{0.895} 
& \textbf{0.1547} 
& 0.0320 
& 0.0274 
& \textbf{0.72609} \\

$\mathcal{W} \rightarrow (\mathcal{S}+\mathcal{I})_{30} \rightarrow \mathcal{R}$ 
& 0.876 
& 0.1654 
& 0.0337 
& \textbf{0.0186} 
& 0.71623 \\

$\mathcal{W} \rightarrow \mathcal{I}_{5} \rightarrow \mathcal{S}_{8} \rightarrow \mathcal{R}$ 
& 0.878 
& 0.1632 
& \textbf{0.0263} 
& 0.0302 
& 0.71703 \\
\bottomrule
\end{tabular*}
\end{table}

\subsection{Transfer Analysis}

\paragraph{Effect of Waymo Initialization.}
Comparing $\mathcal{R}$ (scratch) with $\mathcal{W} \rightarrow \mathcal{R}$ demonstrates the impact of large-scale urban pretraining.  
Waymo initialization consistently improves NDS, indicating that generic object representations transfer to the racing domain despite differences in scene structure.

\paragraph{Effect of Simulator Pretraining.}
Pure simulator pretraining ($\mathcal{W} \rightarrow \mathcal{S}_{10} \rightarrow \mathcal{R}$) improves performance compared to scratch training, but does not always outperform direct finetuning from Waymo.  
This suggests that synthetic racing data alone does not fully bridge the domain gap.

\paragraph{Effect of Indy as Intermediate Domain.}
Pretraining on real racing data from Indy ($\mathcal{W} \rightarrow \mathcal{I}_{10} \rightarrow \mathcal{R}$) yields stronger performance gains than simulator-only pretraining.  
This indicates that domain similarity plays a critical role in high-speed racing perception.

\paragraph{Multi-Domain Pretraining.}
Joint pretraining on $\mathcal{S} \cup \mathcal{I}$ further stabilizes performance, suggesting complementary benefits of synthetic scale and real racing geometry.

Overall, the strongest results are achieved when combining urban pretraining with intermediate racing-domain adaptation before final finetuning on A2RL Real.

\subsection{Trajectory Prediction Results}

\input{sections/5.2_prediction_results}

Table~\ref{tab:prediction_results} summarizes ADE/FDE on train, validation, in-domain test, and $\mathcal{R}$ test for the three $\mathcal{I}$ training setups ($\mathcal{I}^{v5}$, $\mathcal{I}^{v3}$, and $\mathcal{I}$). 
Following our cross-domain protocol, the in-domain test for $\mathcal{I}^{v3}$ is $\mathcal{I}^{v5}$, and the in-domain test for $\mathcal{I}^{v5}$ is $\mathcal{I}^{v3}$.

As shown in Table~\ref{tab:prediction_results}, the $\mathcal{I}^{v5}$ setup provides the strongest overall generalization (best in-domain test and best $\mathcal{R}$ test), while the $\mathcal{I}$ setup achieves the lowest train/validation error but does not provide the best test-time transfer. 
The $\mathcal{I}^{v3}$ setup shows the weakest transfer to $\mathcal{R}$.

Table~\ref{tab:indy_split_difficulty} provides split-difficulty context for these trends. 
In particular, higher turning and acceleration extremes are aligned with higher ADE/FDE, which is consistent with the weaker performance on harder test splits. 
This supports the interpretation that split composition is an important factor in addition to model configuration.

Table~\ref{tab:a2rl_vs_indy} compares $\mathcal{R}$-trained and $\mathcal{I}$-trained models on $\mathcal{R}$. 
The best $\mathcal{I}$ model yields a small ADE gain and a larger FDE gain over the $\mathcal{R}$-trained baseline, indicating improved long-horizon transfer in this setup. 
A likely contributing factor is the relatively limited size of the $\mathcal{R}$ training split.
\begin{table}[t]
\centering
\caption{$\mathcal{I}$ split difficulty (p95) and best trajectory prediction performance (lower is better).}
\label{tab:indy_split_difficulty}
\scriptsize
\setlength{\tabcolsep}{4.5pt}
\renewcommand{\arraystretch}{1.05}
\begin{tabular}{llcccc}
\toprule
Dataset & Split & \#Traj & Accel \(p95\) & Turn \(p95\) (rad) & Best ADE / FDE \\
\midrule
\multirow{3}{*}{$\mathcal{I}$} 
& Train      & 21848 & 0.337 & 0.013 & 0.288 / 0.594 \\
& Validation & 4611  & 0.335 & 0.015 & 0.565 / 1.166 \\
& Test       & 4529  & 2.906 & 2.051 & 1.658 / 3.328 \\
\midrule
\multirow{3}{*}{$\mathcal{I}^{v3}$} 
& Train      & 12259 & 0.349 & 0.013 & 0.512 / 0.888 \\
& Validation & 2959  & 3.031 & 2.135 & 1.556 / 2.988 \\
& Test\(^*\) & 15828 & 0.334 & 0.014 & 0.724 / 1.364 \\
\midrule
\multirow{3}{*}{$\mathcal{I}^{v5}$} 
& Train      & 12668 & 0.328 & 0.014 & 0.528 / 0.924 \\
& Validation & 3130  & 0.368 & 0.016 & 1.131 / 2.233 \\
& Test\(^*\) & 15247 & 0.373 & 0.015 & 0.801 / 1.545 \\
\bottomrule
\end{tabular}
\begin{flushleft}
\footnotesize \(^*\)For $\mathcal{I}^{v3}$ and $\mathcal{I}^{v5}$, the in-domain test split is the complementary $\mathcal{I}$ subset.
\end{flushleft}
\end{table}
\subsection{Qualitative Results}
\begin{figure}[t]
\centering
\includegraphics[width=\columnwidth]{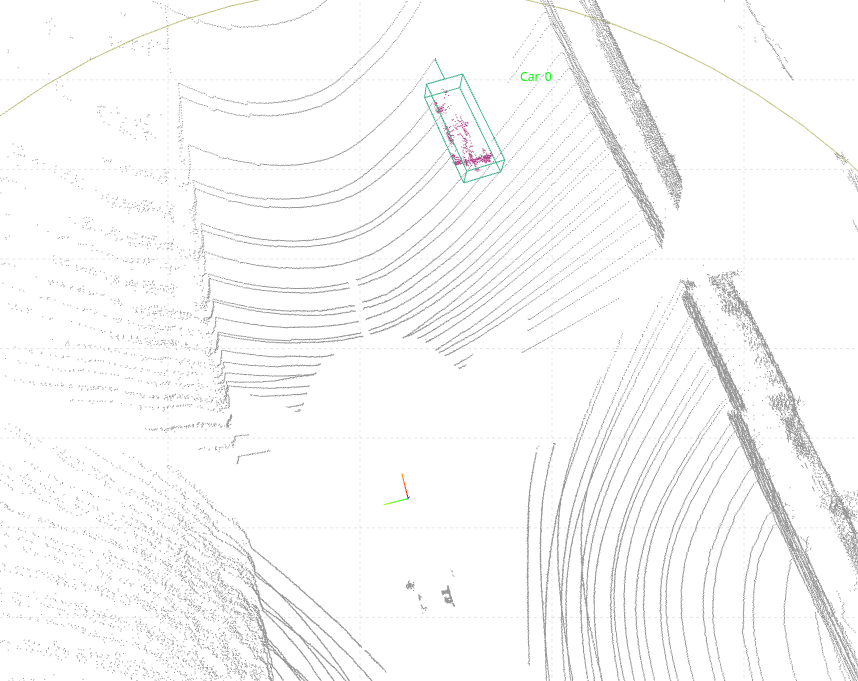}
\caption{LiDAR point cloud visualization of 3D detection result on A2RL Real.}
\label{fig:qualitative}
\end{figure}
Figure~\ref{fig:qualitative} illustrates representative detection outputs on A2RL Real under high-speed overtaking scenarios.

\begin{figure}[t]
\centering
\begin{minipage}[t]{0.49\columnwidth}
    \centering
    \includegraphics[width=\linewidth]{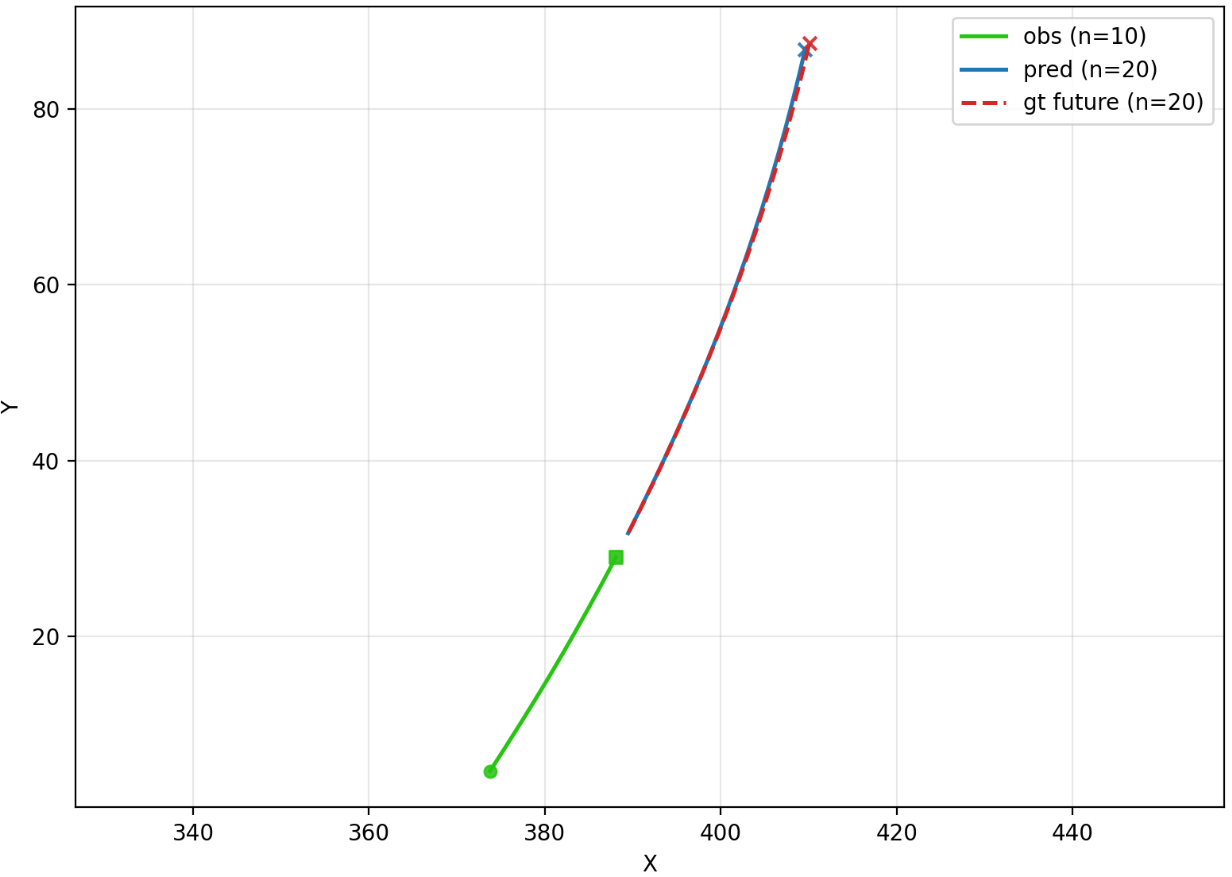}
\end{minipage}\hfill
\begin{minipage}[t]{0.49\columnwidth}
    \centering
    \includegraphics[width=\linewidth]{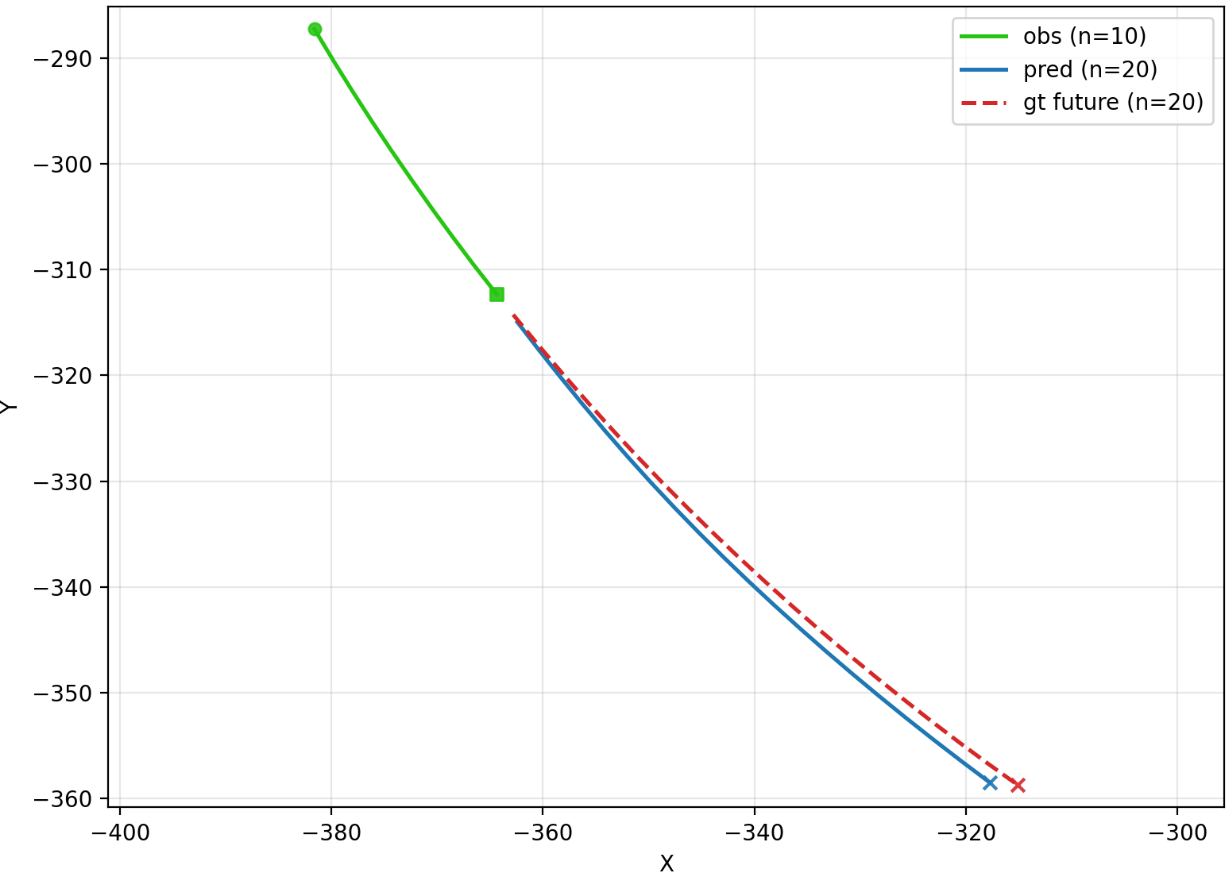}
\end{minipage}
\caption{Representative trajectory prediction examples from the  $\mathcal{I}^{v5}$ training setup, showing observed history and predicted future trajectories.}

\label{fig:qualitative_pred}
\end{figure}

Figure~\ref{fig:qualitative_pred} shows two representative prediction outputs under high-speed racing dynamics.

%% file: sections/5.2_prediction_results.tex
\begin{table}[t]
\centering
\caption{Trajectory prediction performance in terms of ADE and FDE (lower is better).}
\label{tab:prediction_results}
\resizebox{\columnwidth}{!}{
\begin{tabular}{lcccccccc}
\toprule
\multirow{2}{*}{Dataset} 
& \multicolumn{2}{c}{Train}
& \multicolumn{2}{c}{Validation}
& \multicolumn{2}{c}{Test}
& \multicolumn{2}{c}{$\mathcal{R}$} \\
\cmidrule(lr){2-3}
\cmidrule(lr){4-5}
\cmidrule(lr){6-7}
\cmidrule(l){8-9}
& ADE $\downarrow$ & FDE $\downarrow$
& ADE $\downarrow$ & FDE $\downarrow$
& ADE $\downarrow$ & FDE $\downarrow$
& ADE $\downarrow$ & FDE $\downarrow$ \\
\midrule
$\mathcal{I}^{v5}$  
& 0.502 & 0.905
& 1.152 & 2.326
& \textbf{0.774} & \textbf{1.511}
& \textbf{0.478} & \textbf{0.947} \\

$\mathcal{I}^{v3}$  
& 0.672 & 1.175
& 1.624 & 3.254
& 0.862 & 1.611
& 1.567 & 4.205 \\

$\mathcal{I}$  
& \textbf{0.378} & \textbf{0.743}
& \textbf{0.621} & \textbf{1.276}
& 1.611 & 3.177
& 0.852 & 1.389 \\

\bottomrule
\end{tabular}
}
\end{table}
\begin{table}[t]
\centering
\caption{Cross-dataset trajectory prediction results on $\mathcal{R}$. 
Models are trained on $\mathcal{R}$ and $\mathcal{I}$ respectively and evaluated on $\mathcal{R}$ using ADE and FDE ($\downarrow$).}
\label{tab:a2rl_vs_indy}
\resizebox{\columnwidth}{!}{
\begin{tabular}{lcccccc}
\toprule
\multirow{2}{*}{Dataset} 
& \multicolumn{2}{c}{Train}
& \multicolumn{2}{c}{Validation}
& \multicolumn{2}{c}{Test} \\
\cmidrule(lr){2-3}
\cmidrule(lr){4-5}
\cmidrule(lr){6-7}
& ADE $\downarrow$ & FDE $\downarrow$
& ADE $\downarrow$ & FDE $\downarrow$
& ADE $\downarrow$ & FDE $\downarrow$ \\
\midrule
$\mathcal{R}$  
& 0.266 & 0.414
& 0.214 & 0.302
& 0.484 & 1.24 \\

$\mathcal{I}$  
& 0.502 & 0.905
& 1.152 & 2.326
& \textbf{0.478} & \textbf{0.947} \\

\bottomrule
\end{tabular}
}
\end{table}

%% file: sections/6_conclusion.tex
\section{Discussion and Conclusion}
\label{sec:conclusion}

This work presented a unified benchmark for 3D detection and trajectory prediction in high-speed autonomous racing, consolidating data from Indy Autonomous Challenge and A2RL under a consistent annotation protocol. By standardizing LiDAR-based annotations and evaluation procedures, we enabled systematic cross-domain transfer analysis between urban driving, simulator-based racing, and real-world racing scenarios.

Our experiments reveal several key findings. First, large-scale urban pretraining (Waymo) significantly improves racing detection performance compared to training from scratch, demonstrating that generic geometric representations transfer across domains. Second, simulator-only pretraining does not fully bridge the domain gap to real racing data. In contrast, pretraining on real racing data (Indy) provides stronger gains, highlighting the importance of domain similarity over dataset scale. Finally, multi-stage and multi-domain pretraining strategies further stabilize performance, suggesting complementary benefits of synthetic diversity and real-world racing geometry.

Despite these improvements, detection performance remains constrained by the limited size of A2RL Real (929 training frames). Furthermore, the current benchmark considers a single semantic class and does not include velocity or attribute annotations, resulting in a reduced NDS formulation. These limitations reflect the early stage of large-scale perception benchmarks in autonomous racing.

For trajectory prediction, the benchmark also exposes a clear split-composition effect: lower train/validation error does not always yield better test-time transfer. Across Indy training setups, the  $\mathcal{I}^{v5}$ configuration shows the strongest overall generalization (both in-domain and on A2RL), while the full-data setup, despite lower in-split errors, is more sensitive to harder test dynamics (e.g., sharper turns and higher acceleration regimes). We also observe that an Indy-trained model can outperform an A2RL-trained baseline on A2RL test, suggesting that motion-distribution coverage is a key factor for cross-domain forecasting quality.

Future work will investigate domain adaptation techniques tailored to high-speed racing, joint detection–prediction modeling, and multi-agent interaction reasoning. Expanding real-world racing datasets and incorporating additional sensing modalities such as radar may further reduce the sim-to-real gap and improve robustness under extreme dynamics.

Overall, this benchmark provides a structured foundation for perception research in autonomous racing and highlights the critical role of domain-aligned pretraining for high-speed robotic autonomy.